\pdfoutput=1

\documentclass[11pt]{article}

\usepackage[final]{acl}

\usepackage{times}
\usepackage{latexsym}

\usepackage[T1]{fontenc}

\usepackage[utf8]{inputenc}

\usepackage{microtype}

\usepackage{inconsolata}

\usepackage{graphicx}

\usepackage{hyperref}
\usepackage{url}
\usepackage{paralist}
\usepackage{wrapfig}
\usepackage{tcolorbox}

\usepackage{booktabs}
\usepackage{tablefootnote}
\usepackage{amsmath}
\usepackage{enumitem}
\usepackage{algpseudocode}
\usepackage{xspace}
\usepackage{subcaption}
\usepackage{multirow}
\usepackage{multicol}
\usepackage{xcolor}
\usepackage{amsthm}
\usepackage{amssymb}

\usepackage{bbm}
\usepackage{tabularx, booktabs}
\newcolumntype{C}[1]{>{\centering\let\newline\\\arraybackslash\hspace{0pt}}m{#1}}
\newcolumntype{R}[1]{>{\raggedleft\let\newline\\\arraybackslash\hspace{0pt}}m{#1}}
\newcolumntype{L}[1]{>{\raggedright\let\newline\\\arraybackslash\hspace{0pt}}m{#1}}
\usepackage{colortbl}

\usepackage[linesnumbered,ruled,vlined]{algorithm2e}
\SetAlgoNlRelativeSize{-1}

\newcommand*\samethanks[1][\value{footnote}]{\footnotemark[#1]}

\usepackage{pifont}

\hypersetup{
     colorlinks   = true,
     citecolor    = blue
}
\newcommand{\bit}{\begin{compactitem}}
\newcommand{\eit}{\end{compactitem}}
\newcommand{\ben}{\begin{compactenum}}
\newcommand{\een}{\end{compactenum}}

\newtheorem{challenge}{Challenge}

\newcommand{\doconcat}[1]{%
  \vcenter{#1\kern.2ex\hbox{$\Vert$}\kern.2ex}}

\newcommand{\hide}[1]{}

\definecolor{textualcolor}{HTML}{0000FF}
\definecolor{relationalcolor}{HTML}{009900}

\newcommand{\ours}[1]{\cellcolor{green!15}{#1}}
\newcommand{\oracle}[1]{\cellcolor{gray!20}{#1}}

\newcommand{\method}{\textsc{HybGRAG}\xspace}

\newcommand{\grag}{GRAG\xspace}
\newcommand{\problem}{HQA\xspace}

\newcommand{\Chone}{Hybrid-Sourcing Question\xspace}
\newcommand{\Chtwo}{Refinement-Required Question\xspace}
\newcommand{\critic}{critic module\xspace}

\newcommand{\stark}{\textsc{STaRK}\xspace}
\newcommand{\crag}{\textsc{CRAG}\xspace}

\newcommand{\myEmph}[1]{{\em #1}}
\newcommand{\agentic}{\myEmph{Agentic}\xspace}
\newcommand{\general}{\myEmph{Adaptive}\xspace}
\newcommand{\interpretable}{\myEmph{Interpretable}\xspace}
\newcommand{\effective}{\myEmph{Effective}\xspace}

\newcommand{\emphasize}[1]{\textbf{\underline{\smash{#1}}}}

\definecolor{darkgreen}{rgb}{0.0, 0.5, 0.0}
\newcommand*\colorcheck[1]{%
  \expandafter\newcommand\csname #1check\endcsname{\textcolor{#1}{\ding{52}}}%
}
\colorcheck{darkgreen}
\colorcheck{black}

\newcommand{\spaceBelowTableCaption}[0]{\vspace{-1mm}}
\newcommand{\spaceBelowLargeTable}[0]{\vspace{-2mm}}
\newcommand{\spaceabovefigurecaption}[0]{\vspace{-3mm}}
\newcommand{\spaceBelowLargeFigure}[0]{\vspace{-3mm}}

\title{\method: Hybrid Retrieval-Augmented Generation on Textual and Relational Knowledge Bases}

\author{
Meng-Chieh Lee\textsuperscript{\textnormal{1,}}\thanks{The work is done while being an intern at Amazon.} , 
Qi Zhu\textsuperscript{\textnormal{2,}}\thanks{Corresponding authors.} , 
Costas Mavromatis\textsuperscript{\textnormal{2}}, 
Zhen Han\textsuperscript{\textnormal{2}}, 
Soji Adeshina\textsuperscript{\textnormal{2}}, \\
\textbf{
Vassilis N. Ioannidis\textsuperscript{\textnormal{2,}}\samethanks,
Huzefa Rangwala\textsuperscript{\textnormal{2}}, 
Christos Faloutsos\textsuperscript{\textnormal{1,2,}}\samethanks
} \\
\textsuperscript{1}Carnegie Mellon University, 
\textsuperscript{2}Amazon \\
\texttt{\{mengchil,christos\}@cs.cmu.edu}, \\
\texttt{\{qzhuamzn,mavrok,zhenhz,adesojia,ivasilei,rhuzefa\}@amazon.com} \\
}

\begin{document}

\maketitle

\begin{abstract}
Given a semi-structured knowledge base (SKB), where text documents are interconnected by relations, how can we effectively retrieve relevant information to answer user questions?
Retrieval-Augmented Generation (RAG) retrieves documents to assist large language models (LLMs) in question answering; while Graph RAG (\grag) uses structured knowledge bases as its knowledge source.
However, many questions require both textual and relational information from SKB — referred to as ``hybrid'' questions — which complicates the retrieval process and underscores the need for a hybrid retrieval method that leverages both information.
In this paper, through our empirical analysis, we identify key insights that show why existing methods may struggle with hybrid question answering (\problem) over SKB. 
Based on these insights, we propose \method for \problem, consisting of a retriever bank and a \critic, with the following advantages:
(1) \textit{\agentic}, it automatically refines the output by incorporating feedback from the \critic, 
(2) \textit{\general}, it solves hybrid questions requiring both textual and relational information with the retriever bank,
(3) \textit{\interpretable}, it justifies decision making with intuitive refinement path, and
(4) \textit{\effective}, it surpasses all baselines on \problem benchmarks.
In experiments on the \stark benchmark, \method achieves significant performance gains, with an average relative improvement in Hit@$1$ of \textit{51\%}.

\end{abstract}

\section{Introduction}
Retrieval-Augmented Generation (RAG) \citep{lewis2020retrieval, guu2020retrieval} enables large language models (LLMs) to access the information from an unstructured document database.
This allows LLMs to address unknown facts and solve Open-Domain Question Answering (ODQA) with additional textual information.
Graph RAG (\grag) has extended this concept by retrieving information from structured knowledge bases, where documents are interconnected by relationships.
Existing \grag methods focus on two directions:
extracting relational information from knowledge graphs (KGs) and leveraging LLMs for Knowledge Base Question Answering (KBQA) \citep{yasunaga2021qa, sun2024thinkongraph, jin2024graph, mavromatis2024gnn}, and building relationships between documents in the database to improve ODQA performance \citep{li2024graphreader, dong2024don, edge2024local}.



\begin{figure*}[t]
\centering
\includegraphics[width=0.98\textwidth]{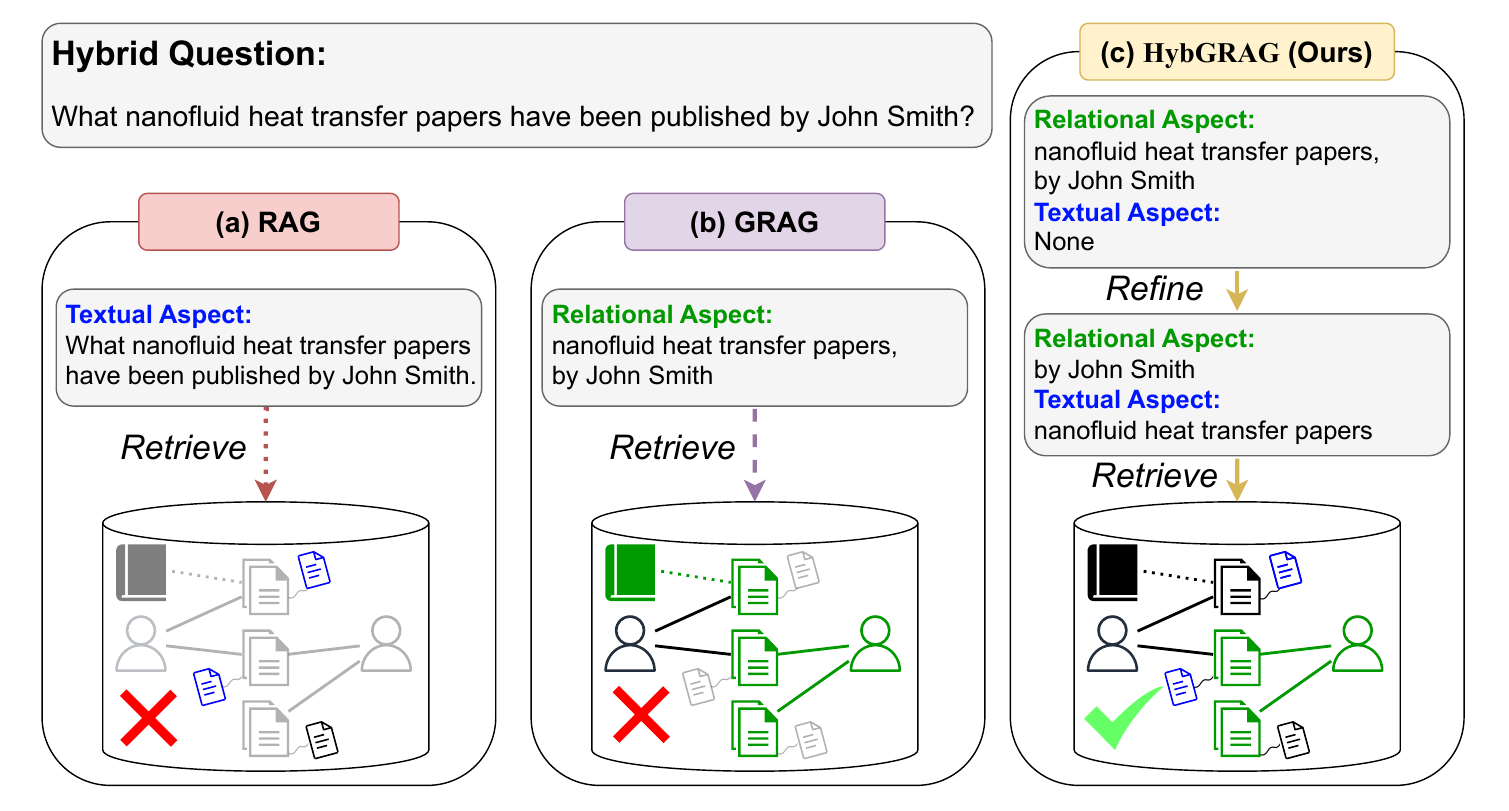}
\spaceBelowLargeFigure
\caption{\emphasize{\method solves hybrid questions in SKB}, which are semi-structured, involving \textcolor{textualcolor}{textual} and \textcolor{relationalcolor}{relational} aspects. 
(a) RAG overlooks the interconnections between documents and does not meet the requirements specified by the relational aspect.
(b) \grag relies solely on the relational aspect and misidentifies the textual aspect as part of the relational one.
(c) \method refines the question routing through self-reflection and successfully retrieves the target document in SKB, addressing both textual and relational aspects.
\label{fig:examples}}
\end{figure*}

Recently, an emerging problem concentrates on ``\textit{hybrid}'' question answering (\problem), where a question requires both relational and textual information to be answered correctly, given a semi-structured knowledge base (SKB) \citep{wu2024stark}.
SKB consists of a structured knowledge base, i.e., knowledge graph (KG), and unstructured text documents, where the text documents are associated with entities of KG.
In Fig.~\ref{fig:examples} top, an example of hybrid questions is given, which involves both the textual aspect (paper topic) and the relational aspect (paper author), and SKBs are the cylinders.

Nevertheless, through our empirical analysis, we uncover two critical insights showing that existing methods that perform RAG or \grag cannot effectively tackle \problem, which requires a synergy between the two retrieval methods.
First, they focus solely on retrieving either textual or relational information.
As shown in Fig.~\ref{fig:examples}(a) and (b), this limitation reduces their applicability when the synergy between the two modalities is required.
Second, in hybrid questions, the aspects required to retrieve different types of information may not be easily distinguishable.
In Fig.~\ref{fig:examples}(c), question routing \citep{li2024retrieval} is performed to identify the aspects of the question.
However, in an unsuccessful routing, confusion between the textual aspect ``nanofluid heat transfer papers'' and the relational aspect ``by John Smith'', leads to incorrect retrieval.



\begin{figure*}[t]
\begin{minipage}[h] {0.49\linewidth}
\captionof{table}{\emphasize{\method matches all properties}, while baselines miss more than one property. \label{tab:salesman}}
\centering{\resizebox{1\columnwidth}{!}{
\begin{tabular}{ l | l | ccc | c }
    \hline
    \multicolumn{2}{c|}{\bf Property} & 
    \rotatebox{90}{Regular RAG} &
    \rotatebox{90}{Think-on-Graph} & 
    \rotatebox{90}{\textsc{AvaTaR}} &
    \rotatebox{90}{\bf \method} \\ 
    \hline
    \multicolumn{2}{l|}{\bf 1. \agentic} & & & \darkgreencheck & \darkgreencheck \\ 
    \hline
    \multirow{3}{*}{\bf 2. \general} & 2.1. Questions in ODQA & \darkgreencheck & & \darkgreencheck & \darkgreencheck \\
    & 2.2. Questions in KBQA & & \darkgreencheck & & \darkgreencheck \\
    & 2.3. Questions in \problem & & & & \darkgreencheck \\ 
    \hline
    \multicolumn{2}{l|}{\bf 3. \interpretable} & \textbf{?} & \darkgreencheck & \darkgreencheck & \darkgreencheck \\ 
    \hline
\end{tabular}
}}
\end{minipage} \hfill
\begin{minipage}[h] {0.49\linewidth}
\centering
\includegraphics[width=0.98\linewidth]{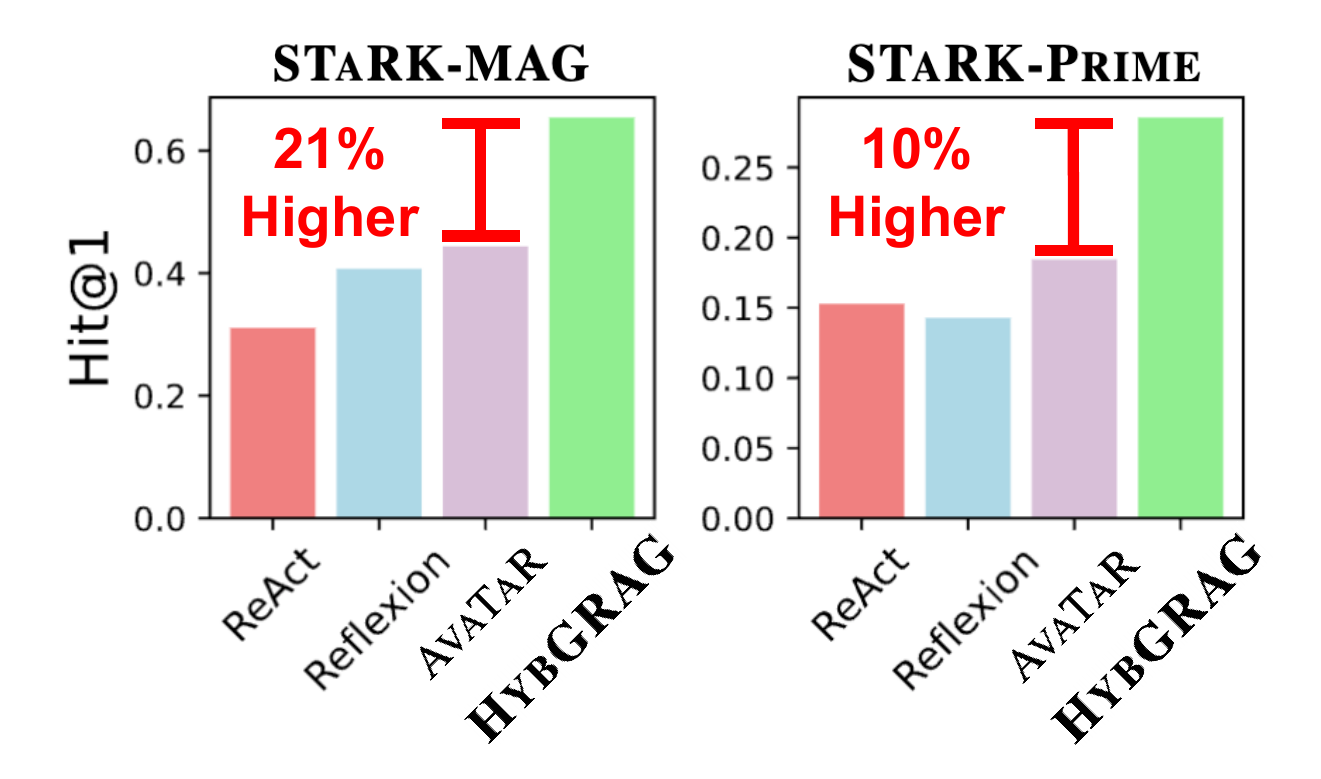}
\spaceabovefigurecaption
\caption{\emphasize{\method wins} in \stark, outperforming baselines by up to $21$\% in Hit@$1$.
\label{fig:crown_jewel}}
\end{minipage}
\end{figure*}

To solve \problem in SKB, we propose \method.
\method handles hybrid questions with a retriever bank, which leverages both textual and relational information simultaneously.
To improve the accuracy of the retrieval, \method performs self-reflection \citep{renze2024self}, which iteratively improves its question routing based on feedback from a carefully designed \critic.
Similarly to chain-of-thought (CoT) \citep{wei2022chain}, which is widely regarded as interpretable, \method's refinement path provides intuitive explanations for the performance improvement.
Last but not least, the framework of \method is designed to be flexible, and can easily be adapted to different problems.
We summarize the contributions of \method as follows:
\ben
    \item \textbf{\agentic}: it automatically refines the question routing with self-reflection;
    \item \textbf{\general}: it solves textual, relational and hybrid questions with a unified framework;
    \item \textbf{\interpretable}: it justifies the decision making with intuitive refinement path; and
    \item \textbf{\effective}: it outperforms all baselines on real-world \problem benchmarks.
\een
As shown in Table~\ref{tab:salesman}, \method is the only work that satisfies all the properties and solves \problem in SKB.
As shown in Fig.~\ref{fig:crown_jewel}, when evaluated on the \problem benchmark \stark, \method outperforms the second-best baseline, achieving relative improvements in Hit@$1$ of \textit{47\%} on \textsc{STaRK-MAG} and \textit{55\%} on \textsc{STaRK-Prime}, respectively.




\section{Proposed Insights: Challenges in \problem}
What new challenges in \problem over SKB remain unsolved by existing methods?
In this section, we first define the problem \problem, and then conduct experiments to uncover two critical insights, laying the foundation for designing our method for \problem.

\subsection{Problem Definition}
A semi-structured knowledge base (SKB) consists of a KG $G=(\mathcal{E}, \mathcal{R})$, where $\mathcal{E}$ and $\mathcal{R}$ represent the sets of entities and relations.
It also includes a set of text documents $\mathcal{D}=\bigcup_{i \in \mathcal{E}}{D_i}$, where $D_i$ is the document associated with entity $i$.
Entity and relation types are denoted by $\mathcal{T}_{E}$ and $\mathcal{T}_{R}$, respectively.
Each hybrid question $q$ in SKB involves semi-structured information, namely, textual and relational information.
We define hybrid question answering (\problem) as follows:
\begin{compactitem}
\item \textbf{Given} a SKB consisting of $G=(\mathcal{E}, \mathcal{R})$ and $\mathcal{D}$, and a hybrid question $q$.
\item \textbf{Retrieve} a set of documents $\mathcal{X} \subseteq \mathcal{E}$, where each document satisfies the requirements specified by the relational and textual aspects of $q$.\looseness=-1
\end{compactitem}

\subsection{C1: \Chone}
To investigate if it is necessary to leverage both textual and relational information to answer hybrid questions, we conduct an experiment to show that text documents and KG contain useful but non-overlapping information.
As a retriever that uses only textual information, vector similarity search (VSS) \citep{karpukhin2020dense} performs retrieval and ranking by comparing the question and documents in the embedding space (``ada-002'');
as a retriever that uses only relational information, Personalized PageRank (PPR) \citep{andersen2006local} performs random walks from the topic entities identified by an LLM (Claude 3 Sonnet) and ranks neighboring entities based on their connectivity in KG.\looseness=-1

In Table~\ref{table:c1}, the text and the graph retrievers have competitive performance.
Interestingly, if an optimal routing always picks the retriever that gives the correct result, the performance is significantly higher, indicating little overlap between the strengths of the text and graph retrievers.
This highlights the importance of a solution to leverage both textual and relational information simultaneously by synergizing these two retrievers.
In Fig.~\ref{fig:examples}(c), we show a hybrid question that requires both textual and relational information to be answered. 
Based on this observation, we uncover the first challenge:
\begin{challenge}[\Chone] \label{ch:c1}
In \problem, there are questions that require both textual and relational information to be answered.
\end{challenge}

\begin{table}[t]
\caption{
    Textual and relational information are both useful to answer hybrid question in \stark-MAG.
}
\spaceBelowTableCaption
\vspace{-1mm}
\centering{\resizebox{0.8\columnwidth}{!}{
\begin{tabular}{ c | c c }
    \toprule
    \textbf{Method} &\textbf{ Hit@$1$} & \textbf{Hit@$5$} \\
    \midrule
    Text Retriever: VSS & 0.2908 & 0.4961 \\
    Graph Retriever: PPR & 0.2533 & 0.5523 \\
    \midrule
    Optimal Routing & 0.4522 & 0.7463 \\
    \bottomrule
\end{tabular}
}}
\label{table:c1}
\spaceBelowLargeTable
\end{table}

\begin{table}[t]
\caption{
    LLMs frequently extracts a subgraph from KG in SKB without target entities in \stark-MAG.
}
\spaceBelowTableCaption
\vspace{-1mm}
\centering{\resizebox{0.8\columnwidth}{!}{
\begin{tabular}{ c | c | c }
    \toprule
    \textbf{\# of Iterations} & \textbf{Feedback Type} & \textbf{Hit Rate} \\
    \midrule
    $1$ & N/A & 0.6769 \\
    $2$ & Simple & 0.7914 \\
    $2$ & Corrective & 0.9231 \\
    \bottomrule
\end{tabular}
}}
\label{table:c2}
\spaceBelowLargeTable
\end{table}


\subsection{C2: \Chtwo}

The success of KBQA often relies on the assumption that the target entities are within an extracted subgraph from KG \citep{lan2022complex}.
Similarly, answering a question in \problem requires extracting a subgraph from KG in SKB.
As hybrid questions involve both textual and relational aspects, they can be challenging for an LLM to comprehend.
To study this, we test if an LLM can extract a subgraph from KG that contains the target entities (hit).
More specifically, an LLM (Claude 3 Sonnet) is prompted to identify the relational aspect in the question, i.e. topic entities and useful relations used to extract the subgraph.
An oracle is used to instruct LLM to perform an extra iteration with feedback if the target entities are not included in the subgraph.\looseness=-1

In Table~\ref{table:c2}, if the result is incorrect, simply prompting LLM to redo the extraction gives a much better hit ratio.
Moreover, if the LLM receives feedback that points out the erroneous part of the extraction (e.g., extracted topic entity is wrong), it significantly improves the result.
This is because in hybrid questions that contain both textual and relational aspects, LLM can falsely identify the textual aspect as the relational one.
In Fig.~\ref{fig:examples} (c), there is an error in retrieving the correct reference from LLM, as it confuses the textual aspect as an entity of type ``field of study'' on the first attempt.
Based on this observation, we uncover the second challenge:
\begin{challenge}[\Chtwo] \label{ch:c2}
In \problem, LLM struggles to distinguish between the textual and relational aspects of the question on the first attempt, necessitating further refinements.
\end{challenge}
\label{sec:discover}

\section{Proposed Method: \method}
To solve \problem, we propose \method, consisting of the \textit{retriever bank} and the \textit{\critic}, to address Challenge~\ref{ch:c1} and Challenge~\ref{ch:c2}, respectively.

\subsection{Retriever Bank (for C1)}
To solve Challenge~\ref{ch:c1}, we propose the retriever bank, composed of multiple retrieval modules and a router.
Given a question $q$, the router determines the selection and usage of the retrieval module, a process known as question routing.
The selected retrieval module then retrieves the top-$K$ references $\mathcal{X}$, as elaborated in the next paragraph.

\begin{table*}[t]
\caption{
    \emphasize{Corrective feedback} of the \critic in \method for \stark.
}
\spaceBelowLargeTable
\centering{\resizebox{1\textwidth}{!}{
\begin{tabular}{ c | l | L{14cm} }
	\toprule
	\textbf{Error Source} & \textbf{Error Type} & \textbf{Feedback} \\
    \midrule
    \multirow{7}{*}{Identification} & Incorrect Entity/Relation & Entity/relation \{name\} is incorrect. Please remove or substitute this entity/relation. \\\cline{2-3}
    & Missing Entity & There is only one entity but there may be more. Please extract one more entity and relation. \\\cline{2-3}
    & No Entity & There is no entity extracted. Please extract at least one entity and one relation. \\\cline{2-3}
    & No Intersection & There is no intersection between the entities. Please remove or substitute one entity and relation. \\\cline{2-3}
    & Incorrect Intersection & There is an intersection between the entities, but the answer is not within it. Please remove or substitute one entity and relation. \\
    \midrule
    Selection & Incorrect Retrieval Module & The retrieved document is incorrect. The current retrieval module may not be helpful to narrow down the search space. \\
	\bottomrule
\end{tabular}
}}
\label{table:criticism_module}
\spaceBelowLargeTable
\end{table*}

\paragraph{Retrieval Modules}
We design two retrieval modules, namely text and hybrid retrieval modules, to retrieve information from text documents and SKB, respectively.
Each retrieval module includes a retriever and a ranker, which offers the flexibility to cover a wide range of questions.

The text retrieval module retrieves documents using similarity search for a given question $q$, such as dense retrieval, which is designed to directly find answers within text documents.
We use VSS between question $q$ and documents $\mathcal{D}$ in the embedding space as both the retriever and the ranker.
This is typically used when nothing can be extracted from the hybrid retrieval module.

The hybrid retrieval module takes the identified topic entities $\hat{\mathcal{E}}$ and useful relations $\hat{\mathcal{R}}$ as input.
It uses a graph retriever to extract entities in the ego-graph of $\hat{\mathcal{E}}$, connected by $\hat{\mathcal{R}}$.
For example, in Fig.~\ref{fig:examples}, $\{\hat{\mathcal{E}}=\{\text{John Smith}\}, \hat{\mathcal{R}}=\{\text{author writes paper}\}\}$ and the graph retriever extracts the entities/papers connected by the path ``John Smith -> author writes paper -> \{paper(s)\}''.
If more than one ego-graph is extracted, their intersection is used as the final result.
Finally, to solve hybrid questions, we propose ranking the documents associated with the extracted entities using a VSS ranker between question $q$ and documents $\mathcal{D}$.
This ensures the synergy between the relational and textual information.

\paragraph{Router}
Given a question $q$, the LLM router performs question routing to determine the selection and usage of the retrieval module.
More specifically, the router first identifies the relational aspect, i.e., topic entities $\hat{\mathcal{E}}$ and useful relations $\hat{\mathcal{R}}$ based on the types of entities $\mathcal{T}_{E}$ and the types of relation $\mathcal{T}_{R}$ using few shot examples \citep{brown2020language}.
The router then makes the selection $s_t$, determining whether to use a text or a hybrid retrieval module.
Identifying $\hat{\mathcal{E}}$ and $\hat{\mathcal{R}}$ before determining $s_t$ improves the quality of $s_t$.
For example, if there is no entity extracted $\hat{\mathcal{E}}=\emptyset$, a text retrieval module is a better option.\looseness=-1



\subsection{Critic Module  (for C2)}
Given a hybrid question $q$, the router is asked to perform question routing, including identifying topic entities $\hat{\mathcal{E}}$ and useful relations $\hat{\mathcal{R}}$.
However, as mentioned in Challenge~\ref{ch:c2}, they may be incorrectly identified in the first iteration.

To solve this, we propose the \critic, which provides feedback to help the router perform better question routing.
Instead of using a single LLM to complete this complicated task, we divide the critic into two parts, an LLM validator $C_{val}$ to validate the correctness of the retrieval $\mathcal{X}$, and an LLM commenter $C_{com}$ to provide feedback $f_t$ if the retrieval is incorrect.
This divide-and-conquer step, similar to previous works \citep{gao2022rarr, asai2024selfrag}, is crucial to our \critic, offering two key advantages:
(1) By breaking a difficult task into two easier ones, we can now leverage pre-trained LLMs to solve them while maintaining good performance.
This resolves the issue when the labels are not available for fine-tuning an LLM critic.
(2) Since the tasks of $C_{val}$ and $C_{com}$ are independent, they can each have their own exclusive contexts, preventing the inclusion of irrelevant information and avoiding the ``lost in the middle'' phenomenon \citep{pmlr-v202-shi23a, liu2024lost}.

\paragraph{Validator}
The LLM validator $C_{val}$ aims to validate if the top references retrieved $\mathcal{X}$ meet the requirements specified by the question $q$, which is a binary classification task.
To improve accuracy, we provide an additional validation context for the validator.
We use the reasoning paths between topic entities and entities in the extracted ego-graph as the validation context, which are used to verify whether the output satisfies certain requirements in the question.
The reasoning paths are verbalized as ``\{topic entity\} \textrightarrow \{useful relation\} \textrightarrow ... \textrightarrow \{useful relation\} \textrightarrow \{neighboring entity\}''.
For example, if a hybrid question asks for a paper (i.e. a document) from a specific author, then the context including the reasoning paths ``\{author\} \textrightarrow \{writes\} \textrightarrow \{paper\}'' is essential for verification.

\paragraph{Commenter}
The LLM commenter $C_{com}$ aims to provide feedback $f$ to help the router refine question routing.
To effectively guide the router, we construct \textit{corrective} feedback that it can easily understand.
In more detail, it points out the error(s) in each action, such as incorrect identification of topic entities, as shown in Table~\ref{table:criticism_module}.
Unlike natural language feedback, which may cause uncertainty or inconsistency depending on the LLM used, our corrective feedback provides clear guidance on how to refine the question routing.
Furthermore, it leverages in-context learning (ICL) to provide sophisticated feedback.
We collect a small number of successful experiences ($\approx 30$) in the training set as examples, with each experience $\{s_{t}, \hat{\mathcal{E}}_t, \hat{\mathcal{R}}_t, f_{t+1}\}$ comprising a pair of router action and feedback, which is verified by ground truth.
During inference, the commenter gives high-quality feedback based on multiple pre-collected examples.

\subsection{Overall Algorithm}
The algorithm of \method is in Algo.~\ref{algo:overall}.
Given a question $q$, in iteration $t$, the router determines $s_{t}$, $\hat{\mathcal{E}}_t$ and $\hat{\mathcal{R}}_t$ to retrieve the references $\mathcal{X}_{t}$ from both $G$ and $\mathcal{D}$ in SKB, or only $\mathcal{D}$, with the selected retrieval module.
The validator $C_{val}$ in the \critic then decides whether to accept $\mathcal{X}_{t}$ as the final answer or reject it.
If $\mathcal{X}_{t}$ is rejected, the commenter $C_{com}$ generates feedback $f_{t+1}$ for the router to assist in refining its action in iteration $t+1$.

\setlength{\textfloatsep}{8pt}
\begin{algorithm}[t]
    \caption{\method \label{algo:overall}}
    \KwIn{Question $q$, a SKB with $G$ and $\mathcal{D}$, Entity Types $\mathcal{T}_{E}$, Relation Types $\mathcal{T}_{R}$, and Maximum Iterations $T$}
    $f_{1}=\texttt{""}$\;
    \For{$t=1, \dots, T$}{
        \textcolor{blue}{\tcc{Retriever Bank}}
        $s_{t}, \hat{\mathcal{E}}_t, \hat{\mathcal{R}}_t=Router(q, \mathcal{T}_{E}, \mathcal{T}_{R}, f_{t})$\;
        \If{$s_{t}$ is hybrid retrieval module}{
            $\mathcal{X}_{t}=HybridRM(q, G, \mathcal{D}, \hat{\mathcal{E}}_t, \hat{\mathcal{R}}_t)$\;
        }
        \Else{
            $\mathcal{X}_{t}=TextRM(q, \mathcal{D})$\;
        }
        \textcolor{blue}{\tcc{Validator}}
        \If{$C_{val}(q, \mathcal{X}_{t})=True$}{
            Return $\mathcal{X}_{t}$\;
        }
        \Else{
            \textcolor{blue}{\tcc{Commenter}}
            $f_{t+1}=C_{com}(q, s_{t}, \hat{\mathcal{E}}_t, \hat{\mathcal{R}}_t)$\;
        }
    }
    Return $\mathcal{X}_{t}$\;
\end{algorithm}

\label{sec:method}

\section{Experiments}
We conduct experiments to answer the following research questions (RQ):
\begin{compactenum}[{RQ}1.]
    \item {\bf Effectiveness:} How well does \method perform in real-world benchmarks?
    \item {\bf Ablation Study:} Are all the design choices in \method necessary?
    \item {\bf Interpretability:} How does \method refine its question routing based on feedback?
\end{compactenum}

\begin{table*}[t]
\caption{
    \emphasize{Retrieval Evaluation on \stark: \method wins.} `*' denotes that only 10\% of the testing questions are evaluated due to the high latency and cost of the methods.
    \colorbox{green!15}{} denotes our proposed method.
}
\spaceBelowTableCaption
\centering{\resizebox{0.98\textwidth}{!}{
\begin{tabular}{ l | cccc | cccc }
	\toprule
	\multirow{2}{*}{\textbf{Method}}
    & \multicolumn{4}{c|}{\textbf{\textsc{STaRK-MAG}}}
    & \multicolumn{4}{c}{\textbf{\textsc{STaRK-Prime}}} \\
    & \textbf{Hit@1} & \textbf{Hit@5} & \textbf{Recall@20} & \textbf{MRR}
    & \textbf{Hit@1} & \textbf{Hit@5} & \textbf{Recall@20} & \textbf{MRR} \\   
    \midrule
    QAGNN
	& 0.1288 & 0.3901
    & 0.4697 & 0.2912
    & 0.0885 & 0.2123
    & 0.2963 & 0.1473 \\
    Think-on-Graph*
	& 0.1316 & 0.1617
    & 0.1130 & 0.1418
    & 0.0607 & 0.1571
    & 0.1307 & 0.1017 \\
    \midrule
    Dense Retriever
	& 0.1051 & 0.3523
    & 0.4211 & 0.2134
    & 0.0446 & 0.2185
    & 0.3013 & 0.1238 \\
    VSS (Text Retrieval Module)
    & 0.2908 & 0.4961
    & 0.4836 & 0.3862
    & 0.1263 & 0.3149
    & 0.3600 & 0.2141 \\
    Multi-VSS
	& 0.2592 & 0.5043
    & \underline{0.5080} & 0.3694
    & 0.1510 & 0.3356
    & 0.3805 & 0.2349 \\
    VSS w/ LLM Reranker*
    & 0.3654 & 0.5317
    & 0.4836 & 0.4415
    & 0.1779 & 0.3690
    & 0.3557 & 0.2627 \\
    \midrule
    ReAct
    & 0.3107 & 0.4949
    & 0.4703 & 0.3925
    & 0.1528 & 0.3195
    & 0.3363 & 0.2276 \\
    Reflexion
    & 0.4071 & 0.5444
    & 0.4955 & 0.4706
    & 0.1428 & 0.3499
    & 0.3852 & 0.2482 \\
    \textsc{AvaTaR}
    & \underline{0.4436} & \underline{0.5966}
    & 0.5063 & \underline{0.5115}
    & \underline{0.1844} & \underline{0.3673}
    & \underline{0.3931} & \underline{0.2673} \\
    \midrule
    Hybrid Retrieval Module (Ours)
    & \ours{0.5028} & \ours{0.5820}
    & \ours{0.5031} & \ours{0.5373}
    & \ours{0.2492} & \ours{0.3274}
    & \ours{0.3366} & \ours{0.2842} \\
    \textbf{\method} (Ours)
	& \ours{\textbf{0.6540}} & \ours{\textbf{0.7531}}
    & \ours{\textbf{0.6570}} & \ours{\textbf{0.6980}}
    & \ours{\textbf{0.2856}} & \ours{\textbf{0.4138}}
    & \ours{\textbf{0.4358}} & \ours{\textbf{0.3449}} \\
    \midrule
    Relative Improvement
	& 47.4\% & 26.2\%
    & 29.3\% & 36.5\%
    & 54.9\% & 12.7\%
    & 10.9\% & 29.0\% \\
    \bottomrule
\end{tabular}
}}
\label{table:expr_stark}
\spaceBelowLargeTable
\end{table*}

\begin{figure*}
\centering
\includegraphics[width=0.9\textwidth]{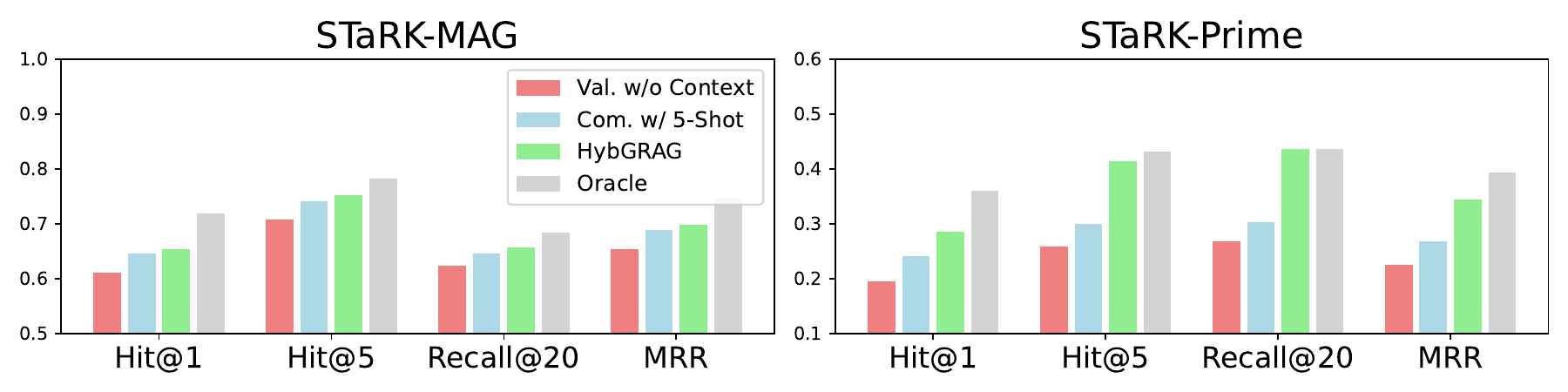}
\spaceabovefigurecaption
\caption{\emphasize{Design choices in \method are necessary} in \stark. 
We compare \method with two variants: a validator without validation context, and a commenter with only 5-shot.
Oracle uses ground truth during inference.
\label{fig:ab_stark}}
\end{figure*}

\begin{figure}
\centering
\includegraphics[width=1\columnwidth]{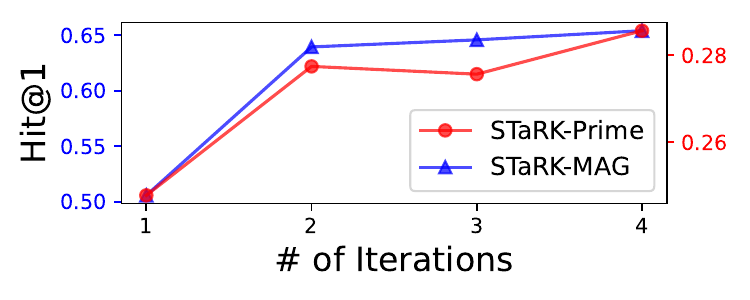}
\spaceabovefigurecaption
\vspace{-3mm}
\caption{\emphasize{\method improves its question routing} thanks to the \critic.
\label{fig:expr_iter_stark}}
\end{figure}

\begin{table}[t]
\captionof{table}{
    \emphasize{\method maintains strong performance} with a less powerful LLM model in \stark-MAG.
}
\spaceBelowTableCaption
\centering{\resizebox{1\columnwidth}{!}{
\begin{tabular}{ l | cccc | c }
\toprule
\textbf{Base Model} & \textbf{Hit@1} & \textbf{Hit@5} & \textbf{Recall@20} & \textbf{MRR} & \textbf{Speedup} \\
\midrule
Claude 3 Haiku & 0.6019 & 0.7084 & 0.6067 & 0.6483 & \textbf{1.96$\times$} \\
Claude 3 Sonnet & \textbf{0.6540} & \textbf{0.7531} & \textbf{0.6570} & \textbf{0.6980} & 1.00$\times$ \\
\bottomrule
\end{tabular}
}}
\label{table:scale}
\end{table}

\begin{table}[t]
\captionof{table}{
    \emphasize{\method performs best with multi-agent}\xspace\emphasize{design} in \stark-MAG.
    ``Router for SR'' baseline performs self-reflection using a single LLM router.
}
\spaceBelowTableCaption
\centering{\resizebox{1\columnwidth}{!}{
\begin{tabular}{ l | r | cccc }
\toprule
\textbf{Method} & \textbf{Setting} & \textbf{Hit@1} & \textbf{Hit@5} & \textbf{Recall@20} & \textbf{MRR} \\
\midrule
Hybrid RM & No-Agent & 0.5028 & 0.5820 & 0.5031 & 0.5373 \\
Router for SR & Single-Agent & 0.6206 & 0.7069 & 0.6187 & 0.6587 \\
\method & Multi-Agent & \textbf{0.6540} & \textbf{0.7531} & \textbf{0.6570} & \textbf{0.6980} \\
\bottomrule
\end{tabular}
}}
\label{table:multiagent}
\end{table}

\paragraph{Benchmarks}
We conduct experiments on two QA benchmarks: \stark\footnote{Due to legal issue, only \textsc{STaRK-MAG} and \textsc{STaRK-Prime} are used in this article.} \citep{wu2024stark}, which serves as the primary evaluation benchmark and focuses on retrieval, and \crag \citep{yang2024crag}, a complementary benchmark to evaluate end-to-end RAG performance.
While \stark focuses on \problem, \crag encompasses both ODQA and KBQA as sub-problems.
Detailed benchmark descriptions are provided in Appx.~\ref{app:data}.

\subsection{Retrieval Evaluation on \stark}
We use the default evaluation metrics provided by \stark, which are Hit@$1$, Hit@$5$, Recall@$20$ and mean reciprocal rank (MRR), to evaluate the performance of the retrieval task.
We compare \method with various baselines, including recent \grag methods (QAGNN~\citep{yasunaga2021qa} and Think-on-Graph~\citep{sun2024thinkongraph}); traditional RAG approaches; and self-reflective LLMs (ReAct~\citep{yao2023react}, Reflexion~\citep{shinn2023reflexion}, and \textsc{AvaTaR}~\citep{wu2024avatar}).
The details of the implementation are in Appx.~\ref{app:rep}.

\subsubsection{Effectiveness (RQ1)}
In Table~\ref{table:expr_stark}, \method outperforms all baselines significantly in both datasets in \stark.
Most baselines are designed to handle ODQA and KBQA, and the results have shown that they cannot handle \problem effectively (Challenge~\ref{ch:c1}).
Our hybrid retrieval module is the second-best performing method, highlighting the importance of designing a synergized retrieval module that uses both textual and relational information simultaneously.
In addition, \method performs significantly better than the hybrid retrieval module, indicating that the extracted entity and relation are frequently incorrect in the first iteration (Challenge~\ref{ch:c2}).
By tackling Challenge~\ref{ch:c1} and~\ref{ch:c2} with our retriever bank and \critic respectively, \method has a significant improvement in performance.

\begin{figure*}[t]
    \centering
    \includegraphics[width=1\linewidth]{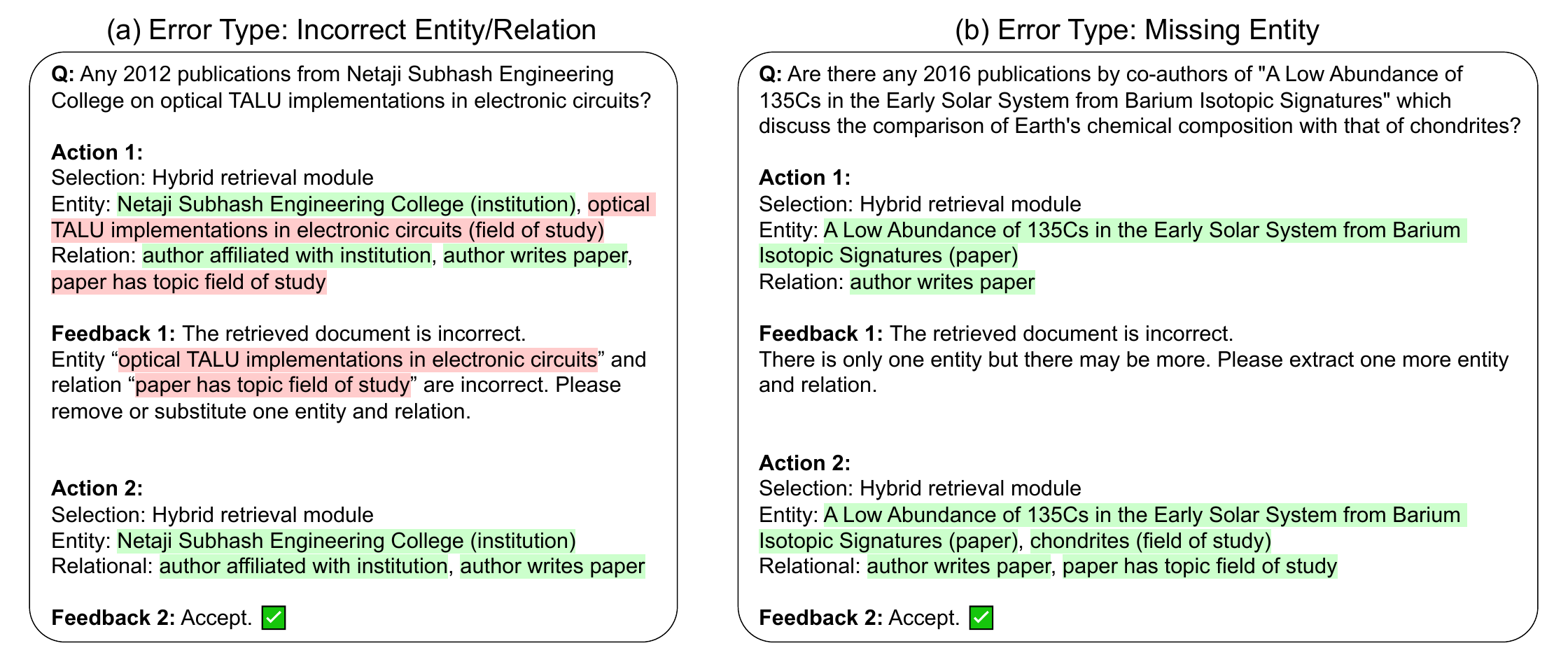}
    \spaceabovefigurecaption
    \vspace{-3mm}
    \caption{\emphasize{\method is interpretable.} In examples from \textsc{STaRK-MAG}, \method successfully refines its entity and relation extraction based on corrective feedback from the \critic.
    }
    \label{fig:explain}
\end{figure*}

\begin{table*}[t]
\caption{
    \emphasize{End-to-End RAG Evaluation on \crag: \method wins.}
    All baselines (except CoT LLM) share our retriever bank, but use different critics to provide feedback.
    \colorbox{green!15}{} denotes our proposed method.
}
\spaceBelowTableCaption
\centering{\resizebox{0.93\textwidth}{!}{
\begin{tabular}{ l | cccr | cccr }
	\toprule
    \multirow{2}{*}{\textbf{Method}}
	    & \multicolumn{4}{c|}{\textbf{Llama 3.1 70B}}
	    & \multicolumn{4}{c}{\textbf{Claude 3 Sonnet}} \\
	     & \textbf{Accuracy}~$\uparrow$ & \textbf{Halluc.}~$\downarrow$ & \textbf{Missing} & \textbf{Score$_{\text{a}}$}~$\uparrow$ & \textbf{Accuracy}~$\uparrow$ & \textbf{Halluc.}~$\downarrow$ & \textbf{Missing} & \textbf{Score$_{\text{a}}$}~$\uparrow$  \\
    \midrule	
    CoT LLM
	& 0.4607 & 0.5026
    & 0.0367 & -0.0419
    & 0.3910 & 0.4052
    & 0.2038 & -0.0142 \\
    Text-Only RAG
	& 0.4105 & 0.3685
    & 0.2210 & 0.0420
    & 0.5034 & 0.3955
    & 0.1011 & 0.1079 \\
    Graph-Only RAG
	& 0.4861 & 0.4442
    & 0.0697 & 0.0419
    & 0.5303 & 0.2974
    & 0.1723 & 0.2329 \\
    Text \& Graph RAG
	& 0.4120 & 0.3790
    & 0.2090 & 0.0330
    & 0.5820 & 0.3416
    & 0.0764 & 0.2404 \\
    \midrule
    ReAct
	& 0.1745 & \textbf{0.2360}
    & 0.5895 & -0.0615
    & 0.4352 & 0.4075
    & 0.1573 & 0.0277 \\
    Corrective RAG
	& 0.4509 & 0.4652
    & 0.0839 & -0.0143
    & 0.4674 & 0.3333
    & 0.1993 & 0.1341 \\
    \midrule
    \textbf{\method} (Ours)
	& \ours{\textbf{0.5206}} & \ours{0.3588}
    & \ours{0.1206} & \ours{\textbf{0.1618}}
    & \ours{\textbf{0.6322}} & \ours{\textbf{0.2959}}
    & \ours{0.0719} & \ours{\textbf{0.3363}} \\
    \bottomrule
\end{tabular}
}}
\label{table:expr_crag}
\end{table*}

\subsubsection{Ablation Study (RQ2)}

\noindent\textbf{Critic Module}
We compare \method variants with a validator without validation context, a commenter with only five shots, and those with oracles.
The oracle has access to the ground truth, which gives the optimal judgement on the correctness of the output and the error type of question routing, if there is any.
In Fig.~\ref{fig:ab_stark}, we show that \method performs the best with all our design choices, approaching the performance of an oracle.

\noindent\textbf{Self-Reflection}
In Fig.~\ref{fig:expr_iter_stark}, we demonstrate that with more self-reflection iterations, the performance of \method improves further.
Performance improves significantly when increasing the number of iterations from $1$ to $2$, where no self-reflection is performed in iteration $1$.
It is also shown that a few iterations are sufficient, as the improvement diminishes over iterations.

\noindent\textbf{Model Size}
Although we do not have access to Claude 3 Opus, we conduct experiments with Claude 3 Haiku, a more cost-efficient but less powerful alternative to Claude 3 Sonnet\footnote{\url{https://www.anthropic.com/news/claude-3-family}}.
In Table~\ref{table:scale}, \method maintains strong performance even with Claude 3 Haiku.
The results also follow the scaling law of LLMs \citep{kaplan2020scaling}.

\noindent\textbf{Multi-Agent Perspective}
Since \method can be interpreted as a multi-agent system, we add a single-agent baseline, which relies on the router to make decisions and provide feedback for self-reflection.
In Table~\ref{table:multiagent}, \method outperforms both single-agent and no-agent baselines.
This highlights that self-reflection is essential for achieving strong performance in \problem, as pointed out in Challenge~\ref{ch:c2}.
Moreover, unlike the plain text feedback generated by the single-agent baseline, the feedback generated by \method more effectively guides the router in refining its decision, thanks to our carefully designed \critic.

\subsubsection{Interpretability (RQ3)}
Fig.~\ref{fig:explain} illustrates examples of the interaction between the router in the retriever bank and the \critic in \textsc{STaRK-MAG}.
In the first iteration of Fig.~\ref{fig:explain}(a), the router misidentifies a ``optical TALU implementations in electronic circuits'' as a topic entity representing the field of study (relational aspect).
Since the ego-graph extracted based on this entity has no intersection with the ego-graph extracted based on ``Netaji Subhash Engineering College'', the \critic recognizes that the former entity has a higher chance of being a textual aspect.
Thus, it gives the feedback to the router, and the router addresses it accordingly.
This refinement path of \method is similar to CoT, making it interpretable and easy for the user to understand.
Examples in \stark-Prime are given in Appx.~\ref{app:prime}.\looseness=-1

\subsection{End-to-End RAG Evaluation on \crag}
We modify \method to adapt to \crag (details in Appx.~\ref{app:rep}).
We use default evaluation metrics with an LLM evaluator to label answers as accurate ($1$), incorrect/hallucination ($-1$), or missing ($0$), yielding Score$_\text{a}$.
We compare \method with CoT LLM, text-only RAG, graph-only RAG, and RAG that concatenates text and graph references.
We include an agentic LLM (ReAct) and a self-reflective LLM (Corrective RAG), both of which share our retriever bank but use different critics.

In Table~\ref{table:expr_crag}, \method outperforms all baselines in \crag.
RAGs with a single retrieval module cannot handle both types of questions.
RAG with a concatenated reference also distracts by irrelevant content in the long reference.
Although our retriever bank is provided, agentic and self-reflective baselines still struggle to refine their actions.
Since ReAct relies on the LLM's ability to think and provide natural language feedback, it often lacks clear guidance.
Without a fine-tuned retrieval evaluator, Corrective RAG cannot effectively identify the usefulness of a reference.

\subsection{Model Cost Analysis}
We report the number of API calls and token consumption (excluding references) for each step in an iteration of \method in Table~\ref{table:api_stark} and~\ref{table:api_crag} for \stark and \crag, respectively.
While most of the token consumption arises from the examples used for ICL, the prompts themselves require very few tokens.
Moreover, since \method uses the chat LLM as the router, the examples for ICL only need to be given once.
Compared to the state-of-the-art baseline \textsc{AvaTaR} in \stark, which requires at least $500$ API calls during training, our hybrid retrieval module achieves a relative improvement $24$\% in Hit@1 with only $2$ API calls, while \method achieves $51$\% with at most $14$ API calls, both without training.\looseness=-1

\begin{table}[t]
\captionof{table}{
    Number of API calls and tokens for \stark.
}
\spaceBelowTableCaption
\centering{\resizebox{1\columnwidth}{!}{
\begin{tabular}{ c | cccc }
\toprule
\textbf{\method} & \textbf{API Call \#} & \textbf{Token \# for} & \textbf{Token \# for} & \textbf{Token \# for} \\
\textbf{Component} & \textbf{per} & \textbf{for} & \textbf{Examples} & \textbf{Examples} \\
 & \textbf{Iteration} & \textbf{Prompts} & \textbf{in MAG} & \textbf{in Prime} \\
\midrule
Router & 2 & 159 & 2709 & 3018 \\
Validator & 1 & 39 & 1383 & 2107 \\
Commenter & 1 & 52 & 1215 & 1583 \\
\bottomrule
\end{tabular}
}}
\label{table:api_stark}
\end{table}

\section{Related Works}
\noindent\textbf{Graph RAG (\grag)}
Various settings have been explored for \grag \citep{peng2024graph}, and can be roughly divided into three directions.
The first focuses on KBQA, taking advantage of the LLM capability \citep{yasunaga2021qa, sun2024thinkongraph, jin2024graph, mavromatis2024gnn}.
The second focuses on ODQA, building relationships between documents to improve retrieval \citep{li2024graphreader, dong2024don, edge2024local}.
The last assumes that a subgraph is given when answering a question \citep{he2024g, hu2024grag}.
In contrast, this paper focuses on solving \problem in SKB, and previous \grag methods are not easily generalized to \problem.

\noindent\textbf{Agentic and Self-Reflective LLMs}
LLM agents \citep{yao2023react,wu2024avatar} facilitate planning in complex reasoning tasks.
Among them, \textsc{AvaTaR} is the most recent, proposing iterative prompt optimization via contrastive reasoning.
However, they may still struggle to generate the correct output on the first attempt.
Self-reflection addresses this limitation by iteratively optimizing the output based on feedback, typically provided by a critic implemented using various approaches: pre-trained LLMs \citep{shinn2023reflexion, madaan2023selfrefine}, external tools \citep{gou2024critic, qiao2024autoact}, or fine-tuned LLMs \citep{paul2024refiner, asai2024selfrag, yan2024corrective}.
Nevertheless, they do not generalize to \problem for two reasons.
First, they lack appropriate retrieval tools and guidance on how to refine retrieval effectively.
Second, in the absence of external tools or labels for fine-tuning, using pre-trained LLMs as critics without careful design results in suboptimal self-evaluation and overly implicit feedback.


\label{sec:related}

\section{Conclusions}
To solve hybrid question answering (\problem), we propose \method, driven by insights from our empirical analysis, which has following advantages:
\ben
    \item \textbf{\agentic}: it refines question routing with self-reflection by our \critic;
    \item \textbf{\general}: it solves textual, relational and hybrid questions by our retriever bank; 
    \item \textbf{\interpretable}: it justifies the decision making with intuitive refinement path; and
    \item \textbf{\effective}: it significantly outperforms all the baselines on \problem benchmarks.
\een
Applied on \stark, \method achieves an average relative improvement $51$\% in Hit@$1$.

\clearpage

\section*{Limitations}
While \method is capable of outperforming existing RAG and \grag methods on \problem, it still has some limitations:
(1) \method uses only the simplest retrieval modules, and various alternatives are not explored.
For example, the ranker in the retrieval modules could be replaced with a cross-encoder ranker, and the retriever in the hybrid retrieval module could use the top-$K$ entities from PPR instead.
(2) \method does not offer significant advantages in terms of domain adaptation. 
In experiments, although \method outperforms baselines, its performance on \stark-Prime is worse than in \stark -MAG, where the academic domain is generally considered less complex than the medicinal domain.
(3) The commentor in \method selects random experiences when performing ICL.
For example, selecting experiences with questions most relevant to the current one may yield better performance.
Although these limitations point out areas for potential improvement, they also present future directions to further enhance the capabilities of \method.

\bibliography{BIB/ref}

\clearpage
\appendix
\section{Appendix: Benchmarks} \label{app:data}

\subsection{\stark} \label{app:datastark}
We use two datasets from the \stark benchmark, \textsc{STaRK-MAG} and \textsc{STaRK-Prime}.
Each dataset contains a knowledge graph (KG) and unstructured documents associated with some types of entities.
The task is to retrieve a set of documents from the database that satisfy the requirements specified in the question.
Noting that the majority of questions are hybrid questions, and there are very few textual questions.
We use the testing set from \stark for evaluation, which contains $2665$ and $2801$ questions for \textsc{STaRK-MAG} and \textsc{STaRK-Prime}, respectively.
The KG of \textsc{STaRK-MAG} is an academic KG, and the one of \textsc{STaRK-Prime} is a precision medicine KG.
Their types of entity and relations are provided in the benchmark.

\subsection{\crag} \label{app:datacrag}
In the \crag benchmark, there are KGs from $5$ different domains that can be utilized to retrieve useful reference.
For each question, a database that includes $50$ retrieved web pages and all $5$ KGs is given, but the answer is not guaranteed to be on the web pages, KGs, or both.
The task is to generate the answer to the question, with or without the help of the retrieved reference.
There are textual and relational questions, covering various question types, such as simple, simple with condition, comparison, and multi-hop.
We use the testing set from \crag for evaluation.
There are $1335$ textual and relation questions, covering various question types, such as simple, comparison, and multi-hop.

\section{Appendix: Experiments} \label{app:expr}

\subsection{Interpretability (RQ3) in \stark-Prime} \label{app:prime}
\begin{figure*}[t]
    \centering
    \includegraphics[width=1\linewidth]{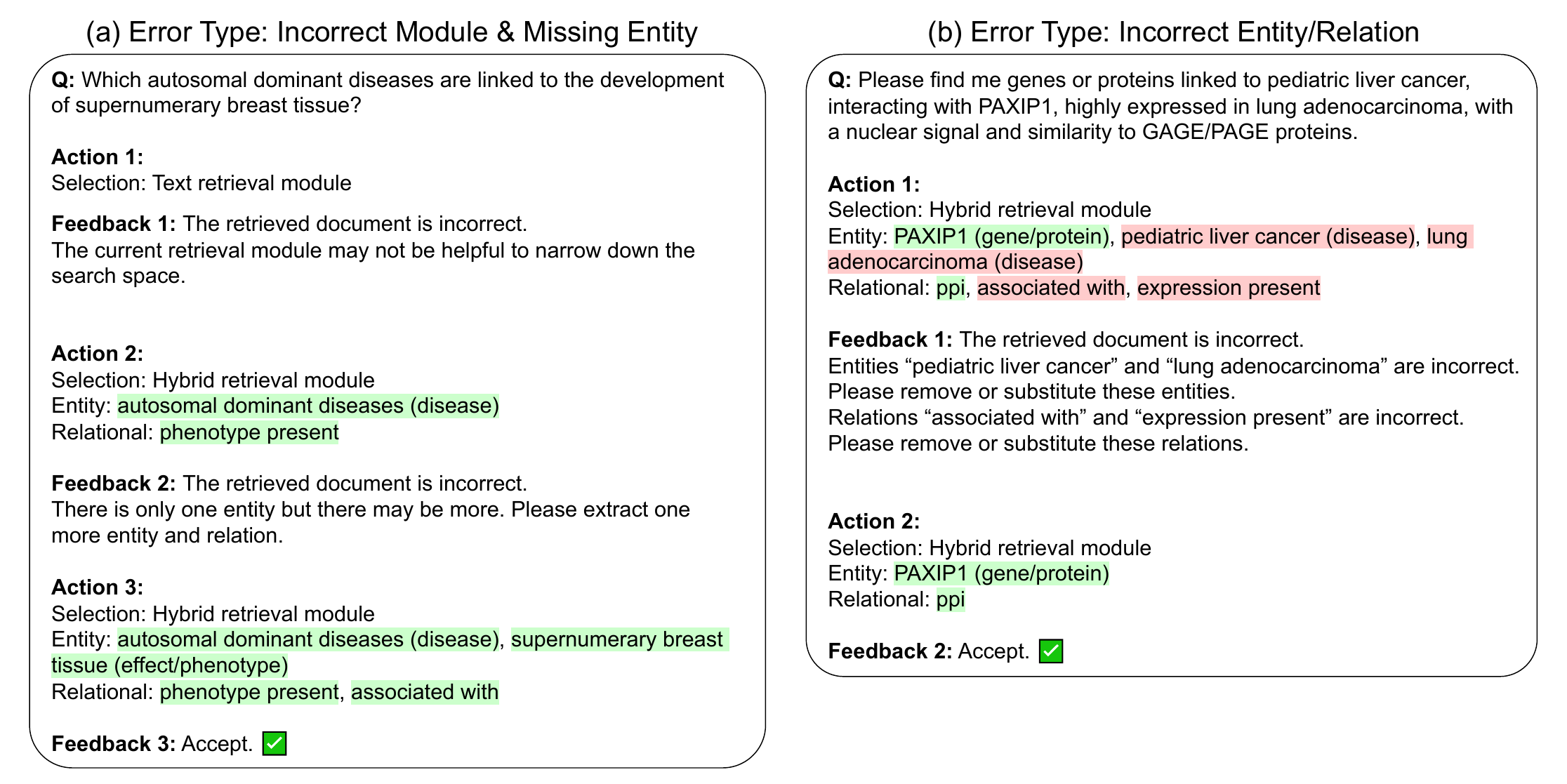}
    \spaceabovefigurecaption
    \caption{\emphasize{\method is interpretable.} In examples from \textsc{STaRK-Prime}, \method successfully refines its entity and relation extraction based on corrective feedback from the \critic.
    }
    \label{fig:explain_prime}
\end{figure*}

Fig.~\ref{fig:explain_prime} shows two examples that \method refines its question routing in \stark-Prime.
In the example of Fig.~\ref{fig:explain_prime}(a), \method selects to use the text retrieval module in the first iteration, and the retrieved document is rejected by the validator.
\method then takes the feedback from the commentor and turns to using the hybrid retrieval module, and refines the extraction of topic entities and useful relations in the next two iterations.

\subsection{Ablation Study on Critic Module} \label{app:critic}
We compare \method variants with validators without validator context, commentors with few or zero shots, and those with oracles.
The oracle has access to the ground truth, which gives the optimal judgement on the correctness of the output and the error type of the action, if there is any.
In Table~\ref{table:expr_stark_critic} and \ref{table:expr_crag_critic}, we show that \method performs the best with all our design choices, approaching the performance of an oracle.

\begin{table*}[t]
\caption{
    \emphasize{The design choices in \method are necessary in \stark.} \colorbox{green!15}{} denotes the settings of \method, and \colorbox{gray!20}{} denotes the baseline that use ground truth during inference.
}
\spaceBelowTableCaption
\centering{\resizebox{1\textwidth}{!}{
\begin{tabular}{ cc | cccc | cccc }
    \toprule
    \multirow{2}{*}{\textbf{Validator}} & \multirow{2}{*}{\textbf{Commentor}}
    & \multicolumn{4}{c|}{\textbf{\textsc{STaRK-MAG}}}
    & \multicolumn{4}{c}{\textbf{\textsc{STaRK-Prime}}} \\
    &  & \textbf{Hit@1} & \textbf{Hit@5} & \textbf{Recall@20} & \textbf{MRR}
    & \textbf{Hit@1} & \textbf{Hit@5} & \textbf{Recall@20} & \textbf{MRR} \\  
    \midrule
	w/o Context & ICL
    & 0.6105 & 0.7073
    & 0.6245 & 0.6541
    & 0.1946 & 0.2592
    & 0.2685 & 0.2251 \\
    w/ Context & $5$-Shot
    & 0.6465 & 0.7407
    & 0.6458 & 0.6884
    & 0.2406 & 0.3006
    & 0.3038 & 0.2676 \\
    \midrule
    w/ Context & ICL
    & \ours{\textbf{0.6540}} & \ours{\textbf{0.7531}}
    & \ours{\textbf{0.6570}} & \ours{\textbf{0.6980}}
    & \ours{\textbf{0.2856}} & \ours{\textbf{0.4138}}
    & \ours{\textbf{0.4358}} & \ours{\textbf{0.3449}} \\
    \midrule
    Oracle & Oracle
    & \oracle{0.7193} & \oracle{0.7824}
    & \oracle{0.6840} & \oracle{0.7479}
    & \oracle{0.3606} & \oracle{0.4320}
    & \oracle{0.4358} & \oracle{0.3932} \\
    \bottomrule
\end{tabular}
}}
\label{table:expr_stark_critic}
\spaceBelowLargeTable
\end{table*}

\begin{table*}[t]
\captionof{table}{
    \emphasize{The design choices in \method are necessary in \crag.} \colorbox{green!15}{} denotes the settings of \method, and \colorbox{gray!20}{} denotes the baseline that use ground truth during inference.
}
\spaceBelowTableCaption
\centering{\resizebox{0.7\textwidth}{!}{
\begin{tabular}{ cc | cccc }
	\toprule
	\textbf{Validator} & \textbf{Commentor} & \textbf{Accuracy}~$\uparrow$ & \textbf{Halluc.}~$\downarrow$ & \textbf{Missing} & \textbf{Score$_{\text{a}}$}~$\uparrow$ \\
    \midrule
    w/o Context & ICL
    & 0.5581 & 0.3461
    & 0.0958 & 0.2120 \\
    w/ Context & $0$-Shot
    & 0.6277 & 0.3004
    & 0.0719 & 0.3273 \\
    \midrule
    w/ Context & ICL
    & \ours{\textbf{0.6322}} & \ours{\textbf{0.2959}}
    & \ours{0.0719} & \ours{\textbf{0.3363}} \\
    \midrule
    Oracle & Oracle
    & \oracle{0.7813} & \oracle{0.1640}
    & \oracle{0.0547} & \oracle{0.6173} \\
	\bottomrule
\end{tabular}
}}
\label{table:expr_crag_critic}
\end{table*}

\section{Appendix: Reproducibility} \label{app:rep}

\subsection{Experimental Details}
All the experiments are conducted on an AWS EC2 P4 instance with NVIDIA A100 GPUs.
Most LLMs are implemented with Amazon Bedrock\footnote{\url{https://aws.amazon.com/bedrock/}}, and Llama 3.1 is implemented with Ollama\footnote{\url{https://github.com/ollama/ollama}}.

\subsubsection{\method Implementation} \label{app:implement}
\paragraph{\stark}
The examples in the prompts are collected from the training set provided by \stark.
We use the default entity and relation types provided by \stark.
The radius of the extracted ego-graph is no more than two.
Four self-reflection iterations have been done.
When extracting the entity name from the question, multiple entities in the knowledge base may have exactly the same name.
In these cases, we select the entity that has the answer in its one-hop neighborhood for disambiguation, since it is not the focus of our paper.
Moreover, these cases rarely happen, where only $3.83$\% and $0.07$\% of questions have this issue in \textsc{STaRK-MAG} and \textsc{STaRK-Prime}, respectively.

\paragraph{\crag}
In the text retrieval module, the web search based on the question is used as the retriever, which is done by \crag ahead of time.
The VSS ranker ranks the web pages based on their similarity to the question in the embedding space.
In this module, we provide an additional choice for the router.
If the output generated based on the current batch of retrieved web pages is rejected by the validator, the router can choose to move on to the next batch in the ranking list.
In \crag, since there is no hybrid question, the hybrid retrieval module is replaced by the graph retrieval module to be prepared for relational questions.
In the graph retrieval module, the retriever extracts the ego-graph connected by the useful relations for each topic entity.
As there is no document associated with entity, the retriever retrieves the reasoning paths from topic entities to entities in the extracted ego-graphs.
Reasoning paths are verbalized as ``\{topic entity\} \textrightarrow \{useful relation\} \textrightarrow ... \textrightarrow \{useful relation\} \textrightarrow \{neighboring entity\}'', and ranked by VSS.

The retrieved reference is used as the validation context to check if it is reliable to answer the question.
The validator takes the output of the generator and the validation context as the input.
As the prompts for the generator and the validator are specialized for different tasks, this allows the validator to offer meaningful validation.
Although the ground truth of the retrieval is not available in \crag, we construct corrective feedback based on the router's action and the evaluation, as shown in Table~\ref{table:criticism_module_crag}.
If the graph retrieval module is used and the evaluation is incorrect, then either the retrieval input (extracted entity and relation or the domain) is incorrect, or selecting graph retrieval module is incorrect;
if the text retrieval module is used and the evaluation is incorrect, then the information in the current batch of documents is considered as not useful to answer the question.

The examples in the prompts are collected from the validation set provided by \crag.
Since the entity and relation types are not given by \crag, and the KGs are only accessible with the provided API, we collect them from the questions in the validation set, as shown in Table~\ref{tab:crag_entity_relation}.
The radius of the extracted ego-graph is no more than two.
Four self-reflection iterations have been done.
A batch contains five web pages.

\begin{table*}[t]
\caption{
    Design of \critic in \method for \crag.
}
\spaceBelowLargeTable
\centering{\resizebox{1\textwidth}{!}{
\begin{tabular}{ c | l | L{10.5cm} }
	\toprule
	\textbf{Error Source} & \textbf{Error Type} & \textbf{Feedback} \\
    \midrule
    \multirow{7}{*}{Input} & Incorrect Question Type & The predicted question type is wrong. Please answer again. Which type is this question? \\\cline{2-3}
    & Incorrect Question Dynamism & The predicted dynamism of the question is wrong. Please answer again. Which dynamism is this question? \\\cline{2-3}
    & Incorrect Question Domain & The predicted domain of the question is wrong. Please answer again. Which domain is this question from? \\\cline{2-3}
    & Incorrect Entity and Relation & The topic entities and useful information extracted from the question are incorrect. Please extract them again. \\
    \midrule
    Selection & Incorrect Retrieval Module & The reference does not contain useful information for solving the question. Should we use knowledge graph as reference source based on newly extracted entity and relation, or use the next batch of text documents as reference source? \\
	\bottomrule
\end{tabular}
}}
\label{table:criticism_module_crag}
\end{table*}

\begin{table*}[t]
\caption{Type of entity and relation in the \crag benchmark.}
\centering{\resizebox{1\textwidth}{!}{
\begin{tabular}{c | l | L{17.5cm}}
\toprule
\textbf{Domain} & \textbf{Type} & \textbf{Content} \\
\midrule
\multirow{5}{*}{Finance} & Entity & company\_name, ticker\_symbol, market\_capitalization, earnings\_per\_share, price\_to\_earnings\_ratio, datetime \\\cline{2-3}
 & Relation & get\_company\_ticker, get\_ticker\_dividends, get\_ticker\_market\_capitalization, get\_ticker\_earnings\_per\_share, get\_ticker\_price\_to\_earnings\_ratio, get\_ticker\_history\_last\_year\_per\_day, get\_ticker\_history\_last\_week\_per\_minute, get\_ticker\_open\_price, get\_ticker\_close\_price, get\_ticker\_high\_price, get\_ticker\_low\_price, get\_ticker\_volume, get\_ticker\_financial\_information \\
\midrule
\multirow{3}{*}{Sports} & Entity & nba\_team\_name, nba\_player, soccer\_team\_name, datetime\_day, datetime\_month, datetime\_year \\\cline{2-3}
 & Relation & get\_nba\_game\_on\_date, get\_soccer\_previous\_games\_on\_date, get\_soccer\_future\_games\_on\_date, get\_nba\_team\_win\_by\_year \\
\midrule
\multirow{5}{*}{Music} & Entity & artist, lifespan, song, release\_date, release\_country, birth\_place, birth\_date, grammy\_award\_count, grammy\_year \\\cline{2-3}
 & Relation & grammy\_get\_best\_artist\_by\_year, grammy\_get\_award\_count\_by\_artist, grammy\_get\_award\_count\_by\_song, grammy\_get\_best\_song\_by\_year, grammy\_get\_award\_date\_by\_artist, grammy\_get\_best\_album\_by\_year, get\_artist\_birth\_place, get\_artist\_birth\_date, get\_members, get\_lifespan, get\_song\_author, get\_song\_release\_country, get\_song\_release\_date, get\_artist\_all\_works \\
\midrule
\multirow{3}{*}{Movie} & Entity & actor, movie, release\_date, original\_title, original\_language, revenue, award\_category \\\cline{2-3}
 & Relation & act\_movie, has\_birthday, has\_character, has\_release\_date, has\_original\_title, has\_original\_language, has\_revenue, has\_crew, has\_job, has\_award\_winner, has\_award\_category \\
\midrule
\multirow{2}{*}{Encyclopedia} & Entity & encyclopedia\_entity \\\cline{2-3}
 & Relation & get\_entity\_information \\
\bottomrule
\end{tabular}
}}
\label{tab:crag_entity_relation}
\end{table*}

\subsubsection{Baseline Implementation}
\paragraph{\stark}
We use ``ada-002'' as the embedding model for dense retrieval and ranking, as used in the paper.
\method uses Claude 3 Sonnet as the base model, while ReAct, Reflexion, \textsc{AvaTaR}, and VSS with LLM reranker use Claude 3 Opus, which is designed to be more powerful than Claude 3 Sonnet\footnote{\url{https://www.anthropic.com/news/claude-3-family}}.
For QAGNN and Dense Retriever, because of the need of training, RoBERTa is used as the base model.
In experiments where the base LLM is not specified, we default to using Claude 3 Sonnet.
We implement Think-on-Graph with their provided code\footnote{\url{https://github.com/GasolSun36/ToG}}, using Claude 3 Sonnet as the base model.
As running the full experiment takes more than a week, we evaluated it with only $10$\% of the testing data, as is done for the LLM reranker in the \stark paper.

\paragraph{\crag}
We use Claude 3 Sonnet as the LLM evaluator, and CoT prompting \citep{wei2022chain} for all generator LLMs.
We use ``BAAI/bge-m3'' \citep{chen2024bge} as the embedding model for dense retrieval and ranking.
ReAct and Corrective RAG share the same backbone with \method, while having different critics.
ReAct has three actions, ``search web'', ``search KG'', and ``extract entity relation domain'', and is given some examples.
The process iterates among action, observation, and thought for four iterations as \method.
While Corrective RAG requires a fine-tuned retrieval evaluator, we implement a version with only a pre-trained LLM.
It starts with the text retrieval module and validates if the retrieved reference is correct, ambiguous, or incorrect.
If incorrect, it uses the graph retrieval module instead.
An final answer is generated based on the reference with CoT prompting.

\subsection{Prompts}

\paragraph{\stark}
The prompt of the router for the first decision making is:
\begin{tcolorbox}
\bfseries
\small
You are a helpful, pattern-following assistant. Given the following question, extract the information from the question as requested.
Rules: 
1. The Relational information must come from the given relational types.
2. Each entity must exactly have one category in the parentheses.\\
$<<<$\{10 examples for entity and relation extraction\}$>>>$
\\\\
Given the following question, based on the entity type and the relation type, extract the topic entities and useful relations from the question.
Entity Type: {$<<<$\{entity types\}$>>>$}\\
Relation Type: {$<<<$\{relation types\}$>>>$}\\
Question: {$<<<$\{question\}$>>>$}
\\\\
Documents are required to answer the given question, and the goal is to search the useful documents. Each entity in the knowledge graph is associated with a document.
Based on the extracted entities and relations, is knowledge graph or text documents helpful to narrow down the search space? You must answer with either of them with no more than two words.
\end{tcolorbox}

The prompt of the router for reflection is:
\begin{tcolorbox}
\bfseries
\small
The retrieved document is incorrect. \\
Feedback: {$<<<$\{feedback on extracted entity and relation\}$>>>$} \\
Question: {$<<<$\{question\}$>>>$}
\\\\
The retrieved document is incorrect. 
Answer again based on newly extracted topic entities and useful relations.
Is knowledge graph or text documents helpful to narrow down the search space? You must answer with either of them with no more than two words.
\end{tcolorbox}

The prompt of the validator is:
\begin{tcolorbox}
\bfseries
\small
You are a helpful, pattern-following assistant.\\
$<<<$\{examples for retrieval validation, 2 for each type of entity\}$>>>$
\\\\
\#\#\# Question: {$<<<$\{question\}$>>>$}\\
\#\#\# Document: {$<<<$\{content of document and reasoning paths\}$>>>$}\\
\#\#\# Task: Is the document aligned with the requirements of the question? Reply with only yes or no.
\end{tcolorbox}

The prompt of the commentor is:
\begin{tcolorbox}
\bfseries
\small
You are a helpful, pattern-following assistant.\\
$<<<$\{30 examples of action and feedback pair\}$>>>$
\\\\
Question: {$<<<$\{question\}$>>>$}\\
Topic Entities: {$<<<$\{extracted entities\}$>>>$}\\
Useful Relations: {$<<<$\{extracted relations\}$>>>$}\\
Please point out the wrong entity or relation extracted from the question, if there is any.
\end{tcolorbox}

\paragraph{\crag}
The prompt of the router for the first decision making is:
\begin{tcolorbox}
\bfseries
\small
You are a helpful, pattern-following assistant.
Given the following question, extract the information from the question as requested.
Rules: 1. Each entity must exactly have one category in the parentheses. 2. Strictly follow the examples.\\
$<<<$\{examples of entity and relation extraction, 5 for each domain\}$>>>$
\\\\
\#\#\# Question Type: simple, simple\_w\_condition, set, comparison, aggregation, multi\_hop, post\_processing, false\_premise.\\
\#\#\# Question: {$<<<$\{question\}$>>>$}\\
\#\#\# Task: Which type is this question? Answer must be one of them.
\\\\
\#\#\# Dynamism: real-time, fast-changing, slow-changing, static.\\
\#\#\# Question: {$<<<$\{question\}$>>>$}\\
\#\#\# Task: Which category of dynamism is this question? Answer with one word and the answer must be one of them.
\\\\ 
\#\#\# Domain: music, movie, finance, sports, encyclopedia.\\
\#\#\# Question: {$<<<$\{question\}$>>>$}\\
\#\#\# Task: Which domain is this question from? Answer with one word and the answer must be one of them.
\\\\
Given the following question, based on the entity type and the relation type, extract the topic entities and useful relations from the question.\\
Entity Type: {$<<<$\{entity types\}$>>>$}\\
Relation Type: {$<<<$\{relation types\}$>>>$}\\
Question: {$<<<$\{question\}$>>>$}
\\\\
\#\#\# Reference Source: knowledge graph, text documents.\\
\#\#\# Question: {$<<<$\{question\}$>>>$}\\
\#\#\# Task: Based on the extracted entity, which reference source is useful to answer the question? You must pick one of them and answer with no more than two words.
\end{tcolorbox}

The prompt of the router for reflection is:
\begin{tcolorbox}
\bfseries
\small
\#\#\# Question Type: simple, simple\_w\_condition, set, comparison, aggregation, multi\_hop, post\_processing, false\_premise.\\
\#\#\# Question: {$<<<$\{question\}$>>>$}\\
\#\#\# Task: The predicted question type is wrong. Please answer again. Which type is this question? Answer with one word and the answer must be one of them.
\\\\
\#\#\# Dynamism: real-time, fast-changing, slow-changing, static.\\
\#\#\# Question: {$<<<$\{question\}$>>>$}\\
\#\#\# Task: The predicted dynamism of the question is wrong. Please answer again. Which dynamism is this question? Answer with one word and the answer must be one of them.
\\\\
\#\#\# Domain: music, movie, finance, sports, encyclopedia.\\
\#\#\# Question: {$<<<$\{question\}$>>>$}\\
\#\#\# Task: The predicted domain of the question is wrong. Please answer again. Which domain is this question from? Answer with one word and the answer must be one of them.
\\\\
The topic entities and useful information extracted from the question are incorrect. Please extract them again.
Given the following question, based on the entity type and the relation type, extract the topic entities and useful relations from the question.\\
Entity Type: {$<<<$\{entity types\}$>>>$}\\
Relation Type: {$<<<$\{relation types\}$>>>$}\\
Question: {$<<<$\{question\}$>>>$}
\\\\
\#\#\# Reference Source: knowledge graph, text documents.\\
\#\#\# Question: {$<<<$\{question\}$>>>$}\\
\#\#\# Task: The answer is incorrect. The reference does not contain useful information for solving the question. Please answer again, should we use knowledge graph as reference source based on newly extracted entity and relation, or use the next batch of text documents as reference source? You must pick one of them and answer with no more than two words.
\end{tcolorbox}

The prompt of the validator is:
\begin{tcolorbox}
\bfseries
\small
\#\#\# Reference: {$<<<$\{reference\}$>>>$}\\
\#\#\# Prediction: {$<<<$\{output of generator\}$>>>$}\\
\#\#\# Question: {$<<<$\{question\}$>>>$}\\
\#\#\# Query Time: {$<<<$\{question time\}$>>>$}\\
\#\#\# Task: The prediction is generated based on the reference. Does the prediction answer the question? Answer with one word, yes or no.
\end{tcolorbox}

The prompt of the commentor is:
\begin{tcolorbox}
\bfseries
\small
You are a helpful, pattern-following assistant.\\
$<<<$\{5 examples of action and feedback pair\}$>>>$
\\\\
\#\#\# Reference Source: {$<<<$\{source\}$>>>$}\\
\#\#\# Question: {$<<<$\{question\}$>>>$}\\
\#\#\# Query Time: {$<<<$\{question time\}$>>>$}\\
\#\#\# Query Type: {$<<<$\{question type\}$>>>$}\\
\#\#\# Query Dynamism: {$<<<$\{dynamism\}$>>>$}\\
\#\#\# Query Domain: {$<<<$\{domain\}$>>>$}\\
\#\#\# Task: Please point out the wrong information about the question (Reference Source, Query Type, Query Dynamism, Query Domain), if there is any. The answer must be one of them.
\end{tcolorbox}

The prompt of the generator is:
\begin{tcolorbox}
\bfseries
\small
You are a helpful, pattern-following assistant.\\
$<<<$\{1 chain-of-though prompt example\}$>>>$
\\\\
\#\#\# Reference: {$<<<$\{reference\}$>>>$}\\
\#\#\# Reference Source: {$<<<$\{source\}$>>>$}\\
\#\#\# Question: {$<<<$\{question\}$>>>$}\\
\#\#\# Query Time: {$<<<$\{question time\}$>>>$}\\
\#\#\# Query Type: {$<<<$\{question type\}$>>>$}\\
\#\#\# Query Dynamism: {$<<<$\{dynamism\}$>>>$}\\
\#\#\# Query Domain: {$<<<$\{domain\}$>>>$}\\
\#\#\# Task: You are given a Question, References and the time when it was asked in the Pacific Time Zone (PT), referred to as Query Time.
The query time is formatted as mm/dd/yyyy, hh:mm:ss PT. The reference may help answer the question.
If the question contains a false premise or assumption, answer “invalid question”.
First, list systematically and in detail all the problems in this problem that need to be solved before we can arrive at the correct answer. Then, solve each sub problem using the answers of previous problems and reach a final solution.
\\\\
What is the final answer?
\end{tcolorbox}

The prompt of the evaluator is:
\begin{tcolorbox}
\bfseries
\small
\#\#\# Question: {$<<<$\{question\}$>>>$}\\
\#\#\# True Answer: {$<<<$\{ground truth answer\}$>>>$}\\
\#\#\# Predicted Answer: {$<<<$\{output of generator\}$>>>$}\\
\#\#\# Task: Based on the question and the true answer, is the predicted answer accurate, incorrect, or missing? The answer must be one of them and is in one word.
\end{tcolorbox}

\begin{table}[h]
\captionof{table}{
    Number of API calls and tokens for \crag.
}
\spaceBelowTableCaption
\centering{\resizebox{0.85\columnwidth}{!}{
\begin{tabular}{ c | ccc }
\toprule
\textbf{\method} & \textbf{API Call \#} & \textbf{Token \# for} & \textbf{Token \# for} \\
\textbf{Component} & \textbf{/ Iteration} & \textbf{Prompts} & \textbf{Examples} \\
\midrule
Router & 4 & 266 & 5752 \\
Validator & 1 & 56 & 0 \\
Commenter & 1 & 78 & 598 \\
Generator & 2 & 168 & 553 \\
\bottomrule
\end{tabular}
}}
\label{table:api_crag}
\end{table}


\end{document}